# A Federated learning model for Electric Energy management using Blockchain Technology

**Muhammad Shoaib Farooq[1,\*], Azeen Ahmed Hayat[4]**
[1]Department of Artificial Intelligence, University of Management and Technology, Lahore, 54000, Pakistan
[4]Institute of Electrical, Electronics and Computer Engineering, University of the Punjab Lahore, Pakistan

Corresponding author: Muhammad Shoaib Farooq (e-mail: Shoaib.farooq@umt.edu.com).

## Abstract
Energy shortfall and electricity load shedding are the main problems for developing countries. The main causes are lack of management in the energy sector and the use of non-renewable energy sources. The improved energy management and use of renewable sources can be significant to resolve energy crisis. It is necessary to increase the use of renewable energy sources (RESs) to meet the increasing energy demand due to high prices of fossil-fuel based energy. Federated learning (FL) is the most emerging technique in the field of artificial intelligence. Federated learning helps to generate global model at server side by ensemble locally trained models at remote edges sites while preserving data privacy. The global model used to predict energy demand to satisfy the needs of consumers. In this article, we have proposed Blockchain based safe distributed ledger technology for transaction of data between prosumer and consumer to ensure their transparency, traceability and security. Furthermore, we have also proposed a Federated learning model to forecast the energy requirements of consumer and prosumer. Moreover, Blockchain has been used to store excess energy data from prosumer for better management of energy between prosumer and grid. Lastly, the experiment results revealed that renewable energy sources have produced better and comparable results to other non-renewable energy resources.

## Introduction
The severe energy crisis in most of the developing countries is putting its fragile economy and unstable national security environment in a difficult situation [1]. They have also forced the closure of hundreds of factories due to high price fossil-fuel based energy at industrial hub [2]. The continuous use and increased demand for energy indicates that energy has become one of the world's most significant issue [4]. Power production, transmission, and distribution businesses have suffered because of the consequently increasing cost of power generating due to high prices of fossil-fuel and increase in percentage of line losses that consequently raised tariffs. This situation motivates to resolve energy crisis using new power-generation sources. The high inflation rate in developing countries uplifts the need of renewable energy otherwise the common man cannot afford the electricity due to high prices [3].

The energy issues that developing countries are currently facing can be resolved by renewable energy sources. Planning and policy makers have been pushed to explore for alternative sources due to the depletion of natural resources and the increasing demand for conventional energy [6]. To satisfy customer demand, developing countries must implement solar photovoltaic cells, which would enhance the country's share of renewable energy [11]. In comparison to conventional energy sources, solar energy technologies (SETs) offer clear environmental benefits, promoting both the sustainable growth of human activities as well as abatement of non - renewable resources. It would also prevent the release of around three gigatons of $CO^2$ annually by coal plants or 20% of the emissions reduction required by 2030 to escape a global warming catastrophe [12]. In parallel, growing privacy risks in popular applications prompted a revision of traditional data training techniques [25]. Fig.1. summarizes the energy produced using multiple sources.

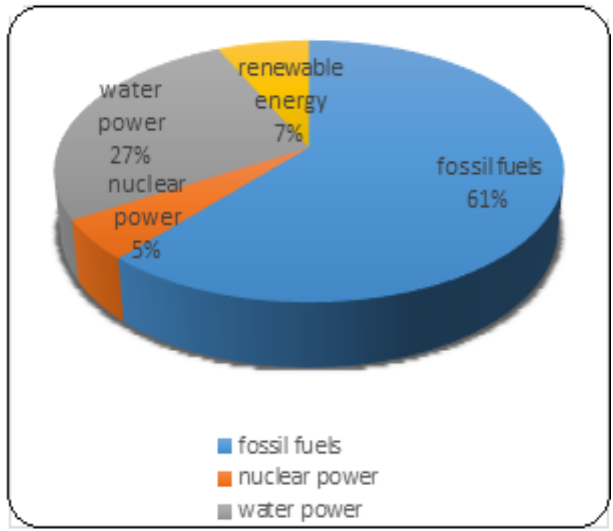 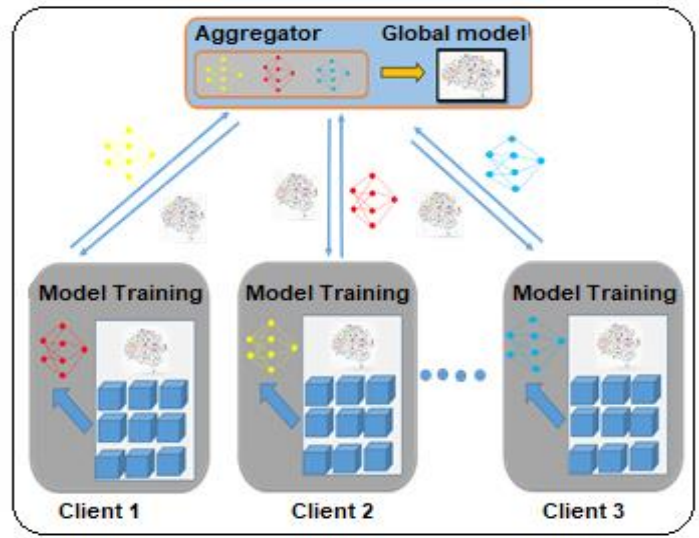

**F*ig.*1** Production capacities per energy source [10]     **Fig. 2** Federated Learning Mechanism

Classical ML requires centralized data training in which data is obtained and the whole training process is carried out at the central server. Despite its high degree of convergence, this training offers a variety of vulnerabilities to the data shared by participants with the primary cloud server. To do so, distributed data training has enforced substantial need for the application of FL. The FL allows users to collaboratively train local models on local data without transmitting any private information to the main cloud server [26]. A distributed machine learning approach called federated learning [5] allows for training on a sizable dataset of decentralized data that is stored on devices like mobile phones. The FL addresses the basic issues of privacy, ownership, and localization of data by taking the more generic approach of "bringing the code to the data, instead of the data to the code". The mechanism of FL is presented in fig. 2. It saves raw data on devices and uses local ML training to reduce data transmission overhead. The learnt and shared models are then federated on a central server to collect and distribute the developed knowledge across participants [7]. Many of the issues preventing the development of the energy Internet can be resolved by applying the technical advantages of the blockchain to it [8]. Distributed ledger technology can improve operational efficiency for utility firms by tracking the chain of custody for grid items. Blockchain offers unique opportunities for renewable energy distribution as well as authenticity monitoring [9].Peer-to-peer energy trading has emerged as the next generation energy management mechanism, allowing multi-level market players to interact more quickly for energy sharing, transferring, exchanging, and trading, with increased engagement of new distributed players and deployment of innovative behind-the-meter activities.

The main goal of this study is to provide a way in which solar energy is generated using solar panels installed in the region. We tried to compensate the effect of other energy sources by implementing solar energy panels (RETs). We have focused on smart management of energy by providing efficient technique to reduce the reasons of energy shortfall. Federated learning and Blockchain technology helped us to propose a model that effectively calculated the potential of energy consumption and generation. How much energy we should have to fulfill the customer demand is the main objective. Blockchain technology helped to trade with the consumers of other region to utilize the surplus energy for reducing energy shortfall. It gives everyone access to a record that is safe, encrypted, clear, accessible, and hard to tamper.

Our proposed model contributes in resolving energy crisis by providing stable model that helps to fulfill consumer and Prosumer requirements. We have implemented a system that contains several of houses with solar panel on rooftops. The consumption and generation data have been used to train model using federated learning, the data is recorded using blockchain technology. Locally Trained model has been sent to server cloud where process of aggregation happened. This process helped to generate a global model which is used to predict the future consumption potential. The proposed system generates an alarm if electricity generation is more than consumption. Blockchain technology also facilitated us in providing successful transaction between Prosumer and grid station. Grid station distributed it to other consumers or client adding monitory to the owner.

The paper is composed of six sections. Section 1 presents an introduction to the energy crisis being faced in developing countries. Related work has been discussed in section 2 showing how various researchers used different techniques in energy prediction and management. Section 3 presents the material and method describing different layers of proposed model.

Experimental analysis and verification have been performed in Section 4. Section 5 presents the results and discussion. Section 6 concludes the research and provide future directions.

## RELATED WORK

With the passage of time, human find different solution to their problems including the problem of deficient non-renewable sources and their harmful effects on the environment. Humans find renewable sources more approachable and cheaper. Recently energy trading and sharing has gathered the attention of researchers [13]. Mostly the already proposed models had used blockchain technology to make trading and sharing encrypted so that the transaction among prosumer and consumer is safe and secure [14]. When we discuss blockchain technology we have to deal with its scalability, security, and decentralization. When comparing base layer approaches, [15] analyzed all of these factors and provided a strategy for enhancing capacity without sacrificing security or decentralization. Although a model to share and trade energy using blockchain has been provided but any solution to calculate future demand of consumers was ignored. [16] suggested a model in which he employed biscotti, a fully decentralized peer-to-peer solution for multi-party ML that coordinates a privacy-preserving ML process between peering clients. Biscotti is a scalable, fault-tolerant, and attack-resistant system that can both protect the privacy of an individual client's update and keep the global model running at scale. although this model also used ML approach but its only deals with privacy issue. However, we gave a solution for future based calculations of consumption and generation with secure transactions among Prosumer and consumer.

Customer service is required for the adoption of renewable energy and the abandonment of conventional fuels for energy generation. Consequently, the consumer participation is critical [17]. Block chain-based peer to peer power trading system that allowed users to contribute to the grid by using renewable energy sources has been provided. Because of the integration of a smart microgrid, the system will be able to function even if it is disconnected from the national grid, allowing it to become self-sufficient. But it has not been able to calculate estimation of future demand of clients. A framework for peer-to-peer energy trading using blockchain is suggested by [18]. It combines four trading systems to accommodate diverse trading preferences of members as well as various electrical generation characteristics and/or consumption. A dual auction Vickrey–Clark–Groves (VCG) procedure has been developed to eliminate any possibility of market power exercise and promote true social welfare by encouraging honest bidding as the main technique of the participants. However, implementation of any technique or algorithm to predict future consumption for the smart management of energy has not been provided.

Since blockchain technology cannot avoid privacy leaking, [19]developed blockchain FL designs. It employs extra data protection techniques. The purpose of this study is to examine how blockchain technology may be used to compensate for flaws in FL. This study does not specify any specific area where it can be used for further development. In cyber-physical systems, scalability and security issues with centralized structure models have created prospects for blockchain-based distributed solutions. A decentralized energy-trading system draws on a variety of sources and effectively coordinates energy to make the most efficient use of available resources. Three blockchain-based energy trading models have been proposed to overcome technical and market constraints, as well as to accelerate the implementation of this disruptive technology [20]. However, the system was not able to provide any estimated data for generation and consumption. [21] proposed machine learning blockchain approach in which smart contracts enable autonomous trade interactions between parties and manage account activity when invoked on blockchain. Based on previous data accumulated on the blockchain, a deep learning-based Gated Recurrent Unit (GRU) model predicts future consumption. Based on the predictions, the K-mean clustering approach is utilized to generate Time of Use (ToU) ranges.

Forecasting solar power generation using semi-asynchronous FL framework while maintaining data privacy and provided a customization strategy to boost model performance further in [22]. But this model only explained FL framework for prediction of generation only. However, in our model the concept blockchain technology has been provided to secure data and for transactions between customers. The issue of accurate Distributed Energy Resources DERs prediction is crucial in distribution grids, not only because distribution systems have limited detection systems, but also because they might have a direct influence on grid-specific functionality such as power balancing [23]. Federated Learning as a distributed machine learning approach has been presented in [23] for DER forecasting through the use of an IoT network, each of which transmitted a model of consumption and generation patterns without disclosing user data. Simulation research with thousand DERs has been implemented to accurately predict customers' privacy along with accurate forecasting. Our houses and schools are powered by modern energy services giving comfort, mobility, and fuel economic activity for production and consumption [24]. We have provided a novel solution to resolve power crisis in which local models have been trained without sharing personal data using federated learning. Two techniques have been applied in our work, first is blockchain that has been used to save the data of consumption and generation.it also produced save transactions between consumers and prosumers. Federated learning has been used to implement training of local models across numerous devices while maintaining data privacy. In our proposed model, federated learning algorithm have been implemented to predict the future energy generation and its demand. Federated

Learning is a potential approach for learning a joint model using all available data. Through this proposed model we have implemented a solution of decreasing energy shortfall. It will help us to shift the energy resources from fossil fuels to RETs that made this model cheaper and more efficient. It also assured that the privacy of every client is the most important part of our model.

## MATERIALS AND METHOD

An essential component of economic activity is energy. The proposed model divides the energy production and consumption system into two broad modules: Consumers and Prosumers. Consumer module comprises of all such house in the selected region that are not contributing in energy production. Whereas prosumers module consists of those houses that generate energy using solar panels, consume electricity as required and trade the electricity that is in access to their needs. Prosumers module divides the flat roofed houses of selected region into four groups as per their covered area. These four types are considered as four clients of the proposed system that have been contributing not only in the production of energy but also in energy trade. Smart meters have been connected outside every house to collect the consumption data, surplus energy data and total production data. Major outliers not considered involved consumers who are unable to purchase electricity due to a lack of finances and prosumers taking control of their extra energy storage, which may be very costly. These costs may have an impact on the prosumer overall profit. Both circumstances may stop potential members from joining this ecosystem. Moreover, ensuring security and privacy of energy trading system is also a big challenge. To overcome these issues, we proposed a unique model that presented a combination of blockchain technology and federated learning to make energy transactions among consumers and prosumers more transparent ensuring security and privacy of electric energy production and consumption system. We envisage the concept of energy sharing as a complement to support energy trading. Energy sharing is the quantity of energy that participants freely trade.

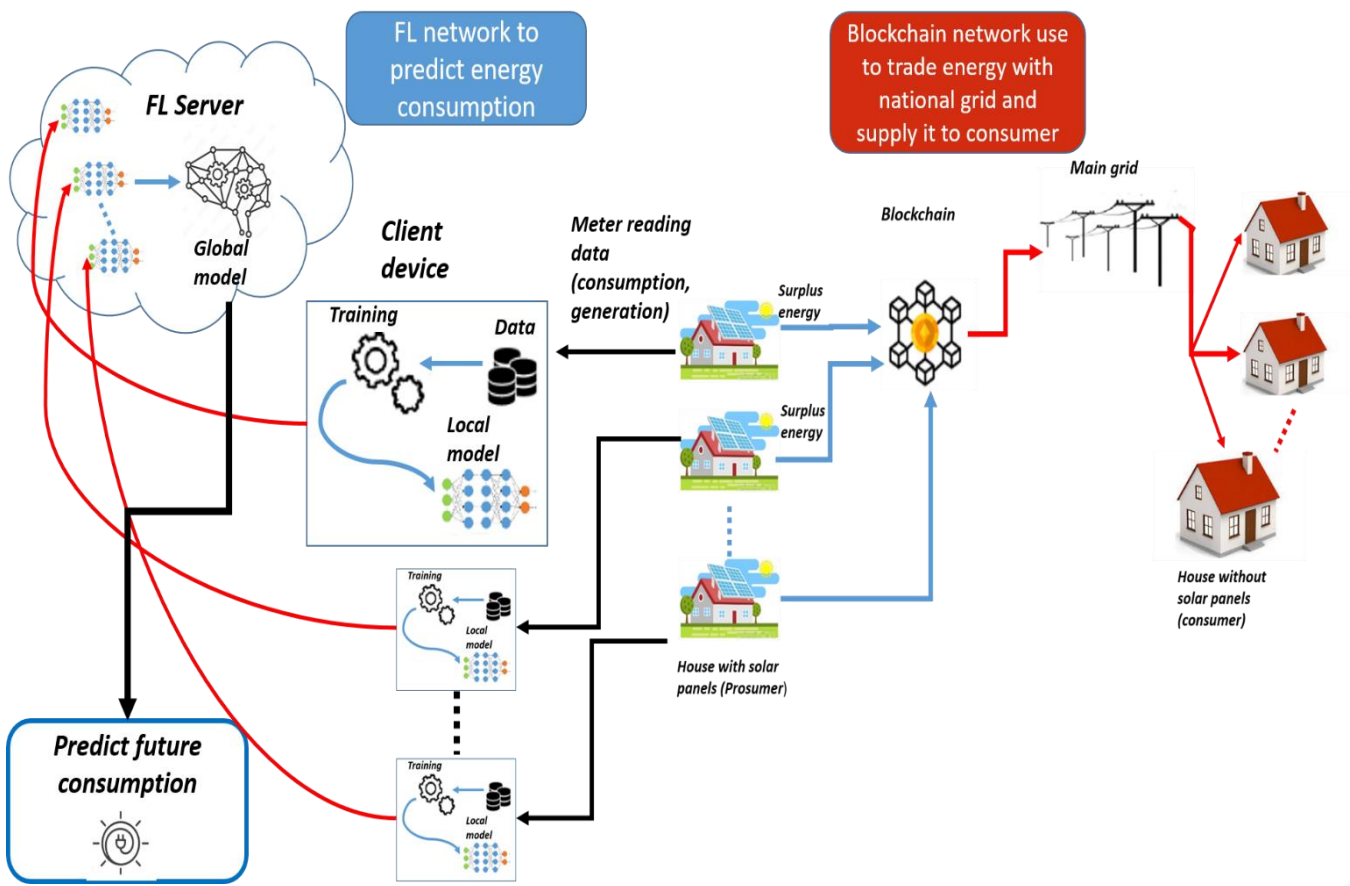

**Fig. 3** Proposed model of our study

The proposed architecture has been divided into two parts: federated section and blockchain section as shown in fig. 4. In federated section the data have been collected using smart meters. Data of electricity produced is transferred to the client server for training local data model. Every client sends model updates to the central server after training local models. The server receives all local updates from all clients and applies federated averaging algorithm (FedGrid) to calculate their mean weight that is further used to train the global model. Hence, a combined global model has been produced using aggregation at central server. Global model anticipates future energy production while keeping in view the consumption rate. The blockchain section calculates total amount of energy consumed in each house and performs save transactions between consumer and Prosumer. To perform energy transactions without the intervention of a third party, the network nodes agree on and execute a desirable and adaptable mutual energy contract. This energy contract is referred to as a Smart Contract (SC) on the blockchain platform, which is an enforceable agreement including specific rules that must be observed by every node of the network. When energy production of any participant exceeds the energy demand, the participant announces that surplus, unused energy will be sold to the grid or adjacent customers.

$$P_s \text{ Production per hour} > P_d \text{ demand per hour}$$

where $P_s$ Production per hour and $P_d$ demand per hour refer to prosumer's energy production and demand on hourly bases. When the energy generated by the participants is insufficient to meet their energy need, they declare themselves as consumers and place an energy request on the network.

$$P_s \text{ Production per hour} < P_d \text{ demand per hour}$$

Prosumers trades energy to the regional grid station through blockchain network. Regional grid station is responsible to supply the surplus energy traded by prosumer to consumer. Consumers purchase electricity from regional grid station. SC enables transparent and reliable energy transfers between network nodes and acts as the governing entity of a decentralized network [27]. SC is immutable, self-executing, code-based, and is kept on the blockchain [28].

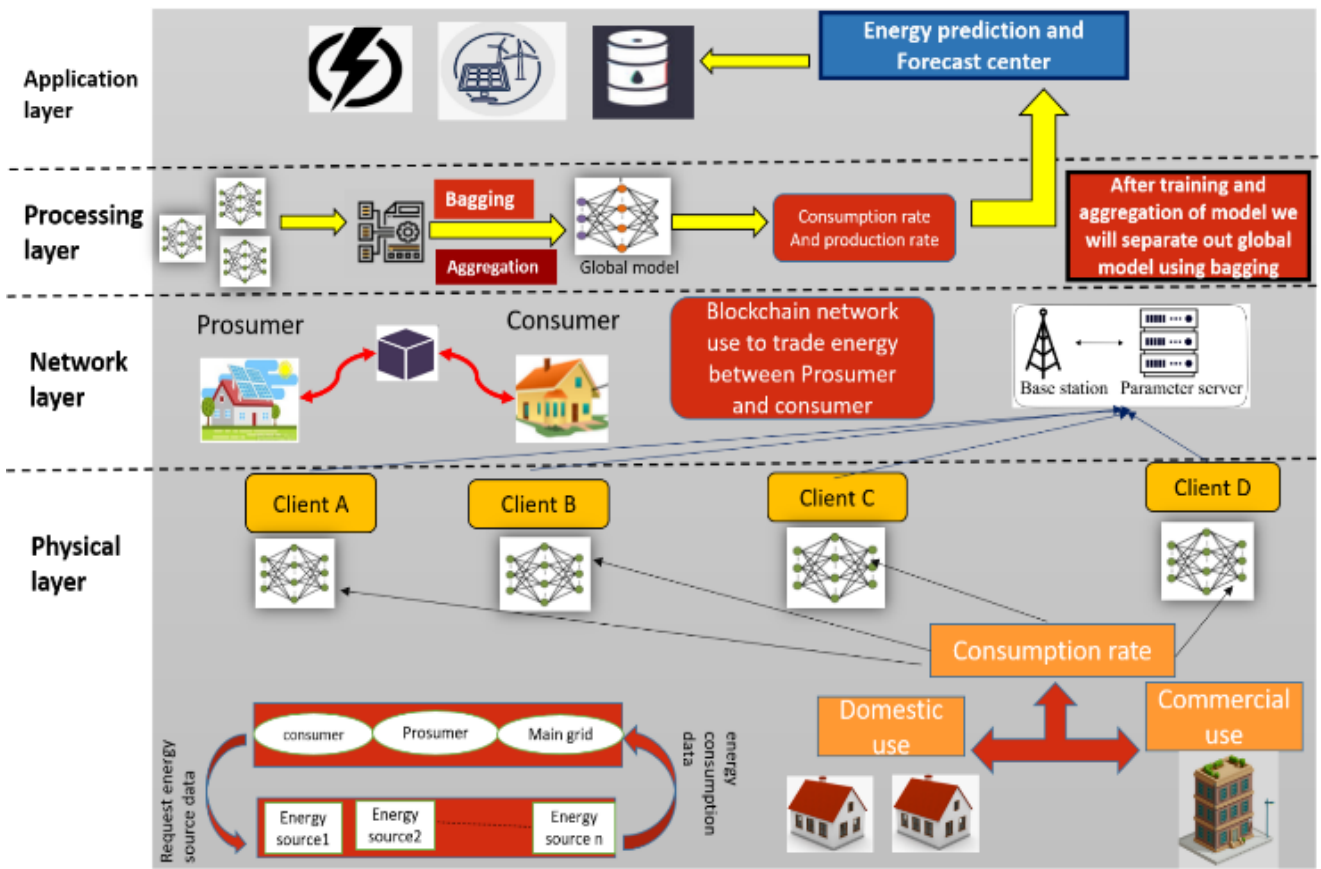

**Fig. 4** Proposed Architecture

The proposed architecture comprises of four layers physical layer, network layer, processing layer, application layer as presented in fig. 5. Data has been collected in physical layer, multiple houses have been grouped as client A, B, C and D for recording real time electricity consumption data of the region according to their covered area. PV panels along with digitalize device and a transmitter have been installed on flat roof tops of every house whereas a smart meter that records total energy consumed per day is connected to each house externally. Digitalizer calculates total energy produced by the house per day and receives total energy consumed per day records from the smart meter to determine energy need. Transmitter device is responsible to collect total energy consumption and total energy production record and transmits these to client servers. Client server also performs some preprocessing on collected dataset like calculating total consumption rate and average consumption rate per house using predefined set of formulas as presented below:

One unit = One kilowatt-hour
Total kilowatt-hour = 1000 Watts $x$ 24 Hours $x$ 30 Days = 720000 watts/hour
Total_units_consumed: 720000/1000...... (k=kilo=1000).
If Total_units_consumed = $\eta$
Avg. cost per unit = $\Lambda$ $.
Total cost = $\eta$ $x$ $\Lambda$ $

Multiple parameters have been identified from the refined dataset for training local data models on client servers and transmitting these local models to the central server through network layer. Network layer works in two folds: transmitting locally trained models to the central server for determining the energy demand through FL and establishing a blockchain network for supplying extra energy produced by prosumers to consumers via blockchain technology that digitalize the transaction of surplus energy from prosumers to fulfil the need of other consumers through regional grid station. Transactions among nodes over a digital network do not require a third party and result in a more cost-effective solution. Grid station has installed multiple poles to supply electricity to consumers. Nodes are users who are connected to the blockchain network. Prosumers claim to sell energy and update their status on the network. Customers that wish to buy electricity make queries on the network. Prosumers and consumers on the blockchain network enter into an agreement under the form of a contract without the requisite of trust.

The processing layer is responsible to perform federated learning-based workflows including training of local data models on clients and training of global data model at the central server. These locally trained models are sent to the central server where the process of bagging is applied to select the most suitable model that will be used to train a global model. The global model has been used to forecast the future demand of every client.

The application layer includes organizations, APIs and grid stations that are interested in energy producing and trading. Energy prediction center and energy trading center are the principal recipients of the forecasts about future energy demand of the region to access the potential prosumers that are willing to trading energy in low prices.

The main goal of this study is to establish a consistent, incredible, and sustainable energy trading and forecast system based on the mutual tradeoff between prosumers, consumers and other agencies.

## EXPERIMENTAL ANAYSIS
Developing countries are facing serios problem of electricity load shading that is significantly affecting industrial and domestic workflows. Regional governments are working hard to find new ways to generate power while considering prospective methods of generating energy by alternative resources. Pakistan being a developing country, needs to produce 25600 MW but currently producing 21000 MW with a gap of 4600 MW between supply and demand [36]. This experiment has been aimed to determine the possibility of solar energy as an alternate source of energy to alleviate load shedding issues from the region of Lahore. A small area of Lahore district has been selected for conducting experiment. Experimental setup has been installed in Wapda employees housing society (WEHS) Lahore presented in fig.5. The WEHS Lahore has been selected due to its suitability for photovoltaic (PV) modules installation. WEHS is a well-planned area with similar structure of houses. A small fraction of rooftops has been analyzed to determine the possible rooftop space for PV panel installation.

The selected region is at 31.4312°N, 74.2444°E and has a total size of 54,900 m$^2$. Lahore is a warm city [29] with high irradiation and a significant potential for PV energy generation.

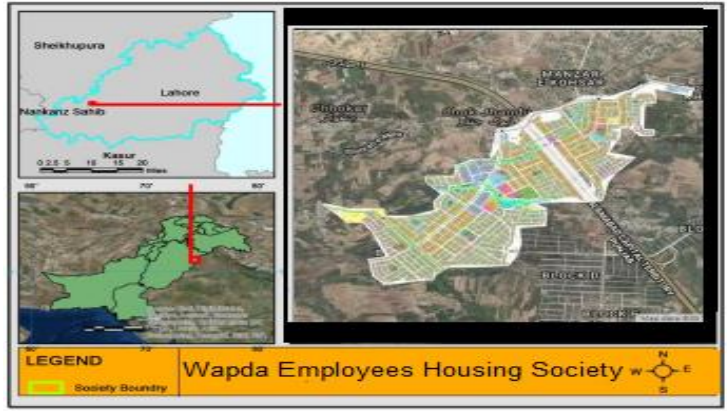

**Fig. 5** Study area

The average temperature remains between 35˚ and 24.6˚ Celsius [30]. May, June, and August were the second hottest months on record, with July being Asia's warmest month on record [31]. In 2021, 705 mm total rainfall has been recorded [32] in the selected region. Least topographic variability makes it very suitable for PV installation in the selected region. The FL, as illustrated in fig. 6, has been applied to compute total solar energy potential on a monthly and yearly basis. To determine the electricity demand for the year 2022, the electricity production and consumption dataset of selected region during January to December 2021 has been recorded.

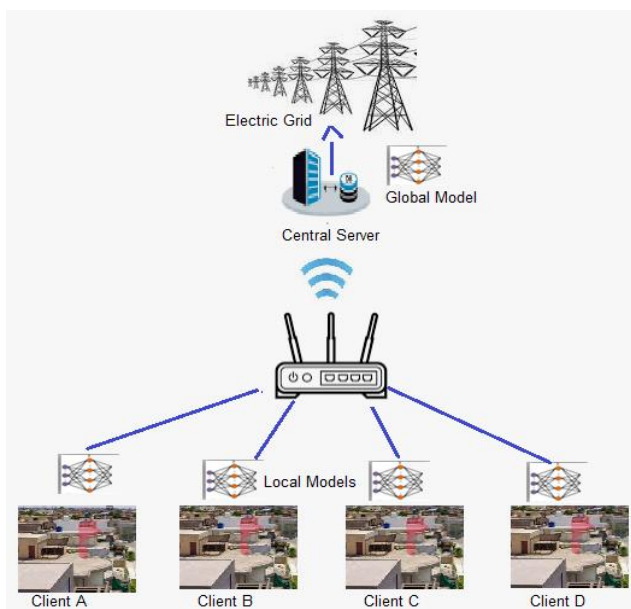
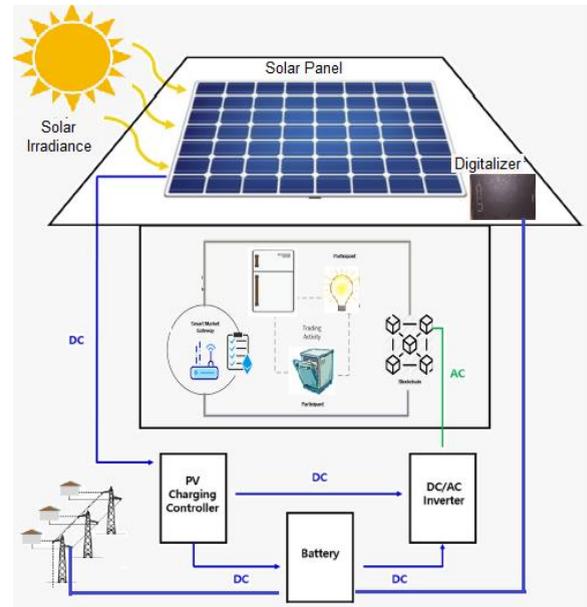

**Fig. 6** Federated Learning model training process     **Fig. 7** Experimental setup of federated blockchain

The calculated production potential of the selected region is 39,613,072 kWh/year. The selected region's monthly total energy usage is 347,140 kWh, which is just 11% of the energy produced by PV solar panels installed on the region's rooftops. The estimated energy generated in the region is about 9 times larger than the specified region's energy consumption; excess energy can be provided to the local or national electricity transmission system. Solar PV energy would be used as a supplement to compensate for the region's energy shortage.

Almost all the roofs are the same height. In the specified location, there are four sorts of houses based on their size. These types are considered as clients in FL model that has been proposed in the Section III. Each client has the same architecture design indicates the same design of all the rooftops. House digitalization criteria has been presented in Table 1. Table 2 summarizes the prosumers and the number of houses. All rooftops have not been digitized. Experimentation has only been performed on houses with same size for digitalization from every client. Few commercial buildings and school rooftop has also been digitized. To avoid shadow problem, mounts of the rooftop have been digitized. Fig. 7 presents the experimental setup of this research. Following criteria has been opted for digitalization.

**TABLE 1** Criteria

| Parameters | Value |
|---|---|
| Direction Of The Building | South, Southeast, Southwest |
| Roof Tilt | Not selected |
| Surface Area | Flat |
| Orientation | Any |
| Shaded Rooftop | Not selected |
| Building Having Hvac System Nor Chimneys | Not selected |

Hence, only such roofs have been selected that receive direct solar radiations. Energy demand for the electrical appliance installed inside the houses has been determined using IoT based smart meters. Total energy consumed is recorded on smart meter from where total demand per house is sent to central server of every client. Client server receives all updates and trains a local model for electric potential demand of the region. In the same way, all clients train their local models and transmit their local models presenting their electricity demand to the central server. Server collects all local models and apply federated averaging algorithm (FedGrid) to train global model presenting total energy demand of the whole region as given below:

```
Algorithm FedGrid
1: Function ServerUpdation
2:  w₀ = 0
3: for t = 1,2… do  // t presents rounds
4: m←max(M. J,1)
5: Sₜ← m   //randomly selected set containing m number of clients
6: for  k ∈ Sₜ do // for every client k in parallel
7: w_{t+1}^{j} ← ClientUpdation(j,wₜ)

8: end_for
9: // privacy preservation strategy
10: w_{t+1}^{j} ← Σ_{j=1}^{J} (n_j/N) w_{t+1}^{j}
11: end_for
12: end_function
13: Function Client Updation (j, ω)
14: // run on client j
15: β← m_j     // Splitting of data with size β of batches
16: for  i = 1 to ε … do // i presents local epoches
17: for batch b ∈ β do
18: w← w − Δ(w)
19: end_for
20: end_for
21: return w    // weight is returned to the server
22: end_Function
```

Energy demand and supply prediction needs big amount of multidimensional data. In the proposed model regional dataset depicting energy consumption has been recorded through different clients. Dataset is processed locally for onsite training of the local data models. Multiple FL parameters are calculated in two steps. Step 1 is on device training of models. In time T = 0 , w₀ along with the mini-batch size (b), learning rate (η), number of trainings (e) is received on the device from server. The receiving device adds its updates and calculates new weight matrix as w₁ = model (x, y, b, e, η). In second step, server collects all local updates from the devices and calculates average weight matrix of 'n' devices as w₁ₙ by applying federated averaging algorithm introduced in this article as FedGrid algorithm.
Total data points (all connected devices) = N

Active data points for $w_{1n}$ weight matrix calculated from Z devices = n

$$w_{1n} = \frac{(n \times w)}{N}$$

At the central server FedGrid algorithm starts with initial weight presented as $w_0$= 0. Now, central server simultaneously communicates with all active local devices and shares model $w_t-1$ with all active clients $S_t$ from total client pool J having participation rate M. One or more training rounds are performed by every active client with learning rate η and model updates are sent back to central server indicating completion of local training. Central server applies fedgrid algorithm on model updates received from connected clients to update and transmit back the basic model towards all active clients . Client takes $T_{local}$ time in completing one round of local training and one complete global round takes $\lambda T_g$ time, where λ is communication delay. Hence

$$T_{global} = \lambda T_g + T_{local}$$

At the same time, smart digitalizes are also working to calculate the amount of electric potential generated by the PV panels. Total amount of electric potential generated by each house is transmitted to the client server. Now, client server will collect data about energy generated by each house and train another local data model presenting the electric potential produced by the client. Local electricity production models are transmitted to central server by each client. Now central server will train global energy production model and transmit it to all clients so that they may add their updates. The study has made use of all freely available data.

Each client records total energy demand of the region and total energy produced in its scope. After calculating electricity production and consumption data, local models are trained and transmitted towards central server. Server receives the energy demand from all clients and applies federated averaging method to calculate the average need of electricity. At the same time, local data models presenting total energy generated through PV panels by every client, is also being transmitted to the central server. Central server is responsible to compare the demand of electricity with total energy produced through solar panels installed in the region. Server generate alerts to the regional grid station when extra power is generated. So that it can be transmitted to other regions through transformers to overcome their electricity shortfall. Federated blockchain based smart grid is presented in fig. 9.

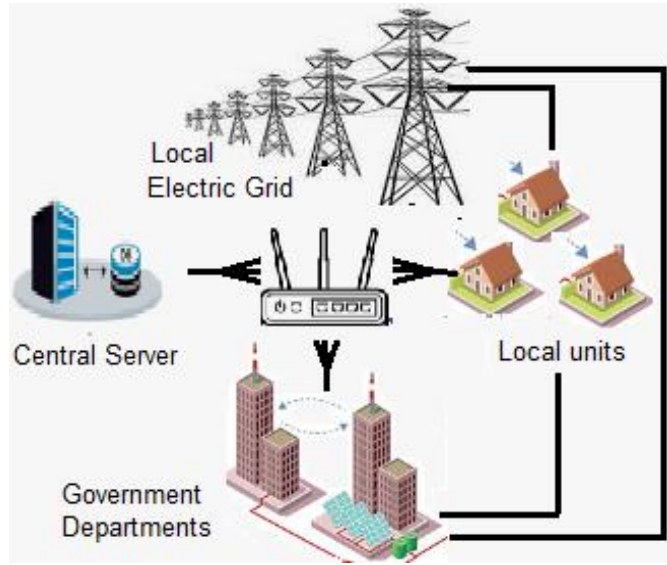

**Fig. 8** Federated blockchain smart grid

At this stage, blockchain technology facilitated the process of demand and supply of electric potential generated and consumed. Major stakeholders involved in this model are People who have installed PV panels on their rooftops, Companies that want to buy the additional electricity produced, local government institutes that manage local grid stations and stable internet WIFI used to transmit all the alerts between all stakeholders.

**TABLE 2** Houses and types

| No. of Houses | House Size |
|---|---|
| 600 | 126.47 $M^2$ |
| 400 | 177.05 $M^2$ |
| 350 | 252.93 $M^2$ |
| 100 | 50586 $M^2$ |
| 06 | 1011 $M^2$ |

Google Earth imagery has been used to digitalize rooftops in the region. On June 18, 2021, high-definition Google Earth images was obtained using Google Earth Pro. WEHS is shown in fig. 10. It presents zero percent cloud cover that is more suitable for digitizing rooftops. As all the roofs in the research region have the same slop and aspect, each meter of roof produces the same amount of energy from solar flux. After digitization total area was calculated for individual building. Table 3 depicts the areas where solar panels can be installed on rooftops. In general, the entire roof is available, but we recommend a small piece of the roof for the intended purpose. House mounts are the most recommended place.

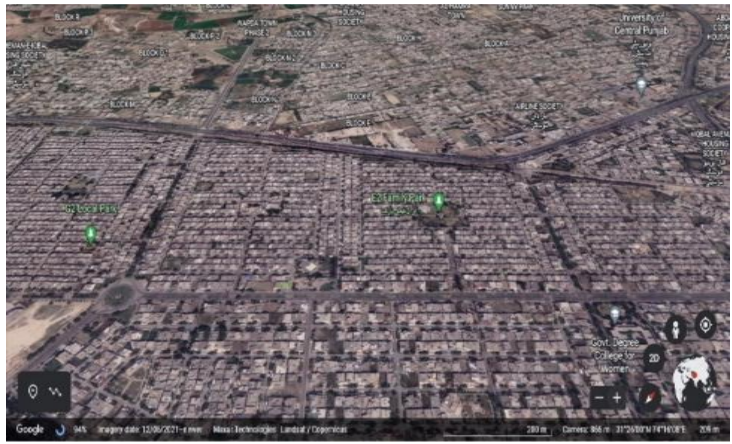

**Fig. 9** Arial view of WEHS Lahore

PV panels have been installed on total 35936 $m^2$ area. Table 4 presents that total electrical power generated by installing PV panels on total area calculated in Table 3.

**TABLE 3** Suitable rooftop area digitalized

| Client | House Size ($M^2$) | Avg area/house ($m^2$) | Total area/ house ($m^2$) | % of total area |
|---|---|---|---|---|
| D | 126.47 | 17 | 10200 | 13.44% |
| C | 177.05 | 28 | 11200 | 15.8% |
| B | 252.93 | 30 | 10500 | 11.9% |
| A | 505.86 | 36 | 3700 | 7.3% |
| Other | 1011 | 56 | 336 | 5.54% |
| Total area of PV installation 35936 $m^2$ | | | | |

It can be observed from Table 4 those different clients are generating different amount of electric potential. Generation of electric potential is directly related to the area of PV installation. Total electric energy generated by digitalization is 37188528 kWh. Table 4 simulates the energy consumed by every house per month. Client D's total power generation would be 718225 kWh/$m^2$ monthly and 8618700 kWh/$m^2$ yearly if all acceptable and projected rooftop space is equipped with solar PV panels. Similarly, Client C generates 527145kWh/$m^2$ per month, Client B produces 475123 kWh/$m^2$ per month, Client A produces 1321286 kWh/$m^2$ per month and 15855432 kWh/$m^2$ would be produced yearly. Table 4 details the monthly as well as yearly potential of each class and total potential of the complete region.

TABLE 4  Solar power production estimate for all houses

| Client | House Size (M$^2$) | Total Estimated roof area | Monthly Potential total area kWh | Yearly potential total area |
|---|---|---|---|---|
| E | Other | 336 | 57265 | 687180 |
| D | 126.47 | 10200 | 718225 | 8618700 |
| C | 177.05 | 11200 | 527145 | 6325740 |
| B | 252.93 | 10500 | 475123 | 5701476 |
| A | 505.86 | 3600 | 1321286 | 15855432 |
| Total | | 35836 | 3099044 | 37188528 |

Table 5 estimates percentage of total household energy usage. Client D's energy usage is 188 kWh per month. Client C's monthly consumption is 270 kWh, Client B's is 310 kWh, and Client A's is 500 kWh.

TABLE 5  Electricity consumption and potential of individual house at WEHS Lahore

| Client | House Size | Avg. Electricity estimate / house / month kWh | |
|---|---|---|---|
| | | Consumption | Electric Potential |
| D | 126.47M$^2$ | 188 | 1187 |
| C | 177.05 M$^2$ | 270 | 1930 |
| B | 252.93 M$^2$ | 310 | 2112 |
| A | 505.86 M$^2$ | 500 | 11909 |

## RESULT AND DISCUSSION

Table 6 compares the total energy consumed by all houses in the selected region with the overall energy produced by PV installation. It can be observed from table 6 that client D and C consume 14% and 13% of the energy produced by them. Client B consumes 16% electric energy from the produced energy. Whereas Client A only consumes only 3 % of self-generated electric energy. In total only 9.2 % of the electric energy generated by installation of PV panels has been consumed indicating the gain of 90.8% of electric potential. Additional electric energy generated in the region can be provided to other areas to overcome electric shortfall.

TABLE 6  Electricity consumption (con) of all houses in selected region [35]

| Client | House Size (M$^2$) | Total Estimated con. per month kWh | Monthly Total Potential kWh | % consumption |
|---|---|---|---|---|
| D | 126.47 | 111,735 | 779,235 | 14% |
| C | 177.05 | 90,321 | 697,786 | 13% |
| B | 252.93 | 95,290 | 595,140 | 16% |
| A | 505.86 | 56,767 | 1,795,341 | 3% |
| Total | | 354,144 | 3,867,502 | 9.2% |

At least same output solar energy potential must be required in the study area for the year of 2023. Recently it is 3,867,502 kWh/m$^2$/year as shown in the table 6. Keeping the current electricity produced as a threshold value with the passage of time due more construction electric potential requirements would increase drastically.

It can be observed from Table 6 that more energy is being produced by Client D. it is due to more number of houses in that category. Houses belonging to the Client A have maximum energy generation potential because of their bigger size or more covered area. Another client E comprising of commercial areas, schools and hospital buildings. High electric potential can be generated by installing PV panels on their rooftop due to more roof top space availability and zero fee.

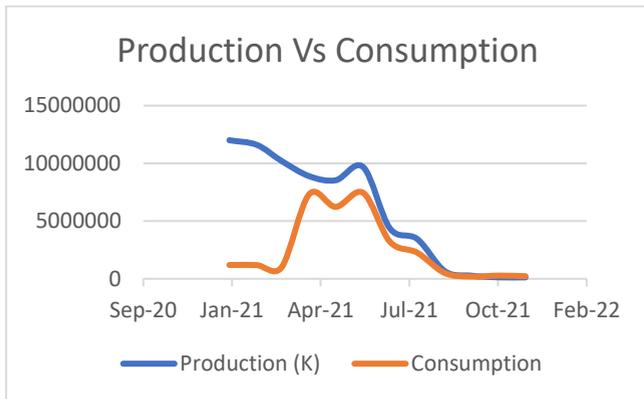
**Fig. 10** Production VS consumption

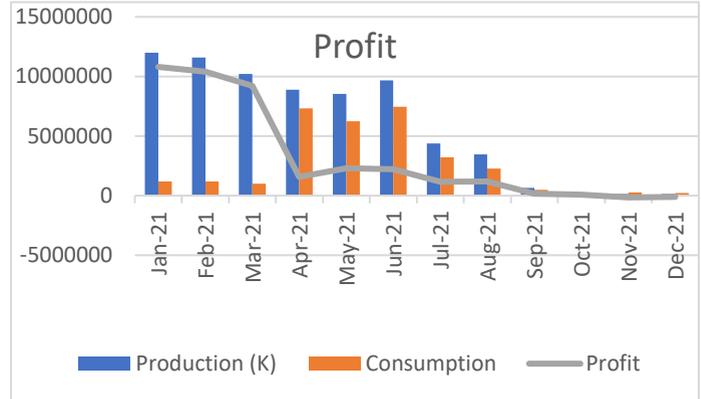
**Fig. 11** Profit Rate

Fig.10 shows that the research region has greater potential for energy generation than demand. The study location is in a region where, from April to July, days are longer owing to the summer season, and energy shortfall is at its greatest. Solar energy generation would be at its greatest due to the summer season. Furthermore, house owners would not only receive lower electricity bills, but they would also receive additional cash from investing in the national grid. In this experiment, very small area has been utilized for PV installation. If the entire roof space was used to generate power, the amount of energy produced from the study area would be enormous.

Fig. 11 presents the gained electricity that has been calculated by comparing the electricity produced and consumption. Even though only a little portion of the roof is chosen, the energy output is extremely great. According to the electric potential scales defined by National Renewable Energy Lab, the research region has a high resource potential.

Fig. 12 presents the position of sun on the selected area during summer and winter seasons. The sun curves have been presented by small diamonds in fig. 13. It can be observed during winters that the sun has low horizon and during summers it has high horizon over the study area. Moreover, maximum amount of solar wave falls in south aspect of the horizon.

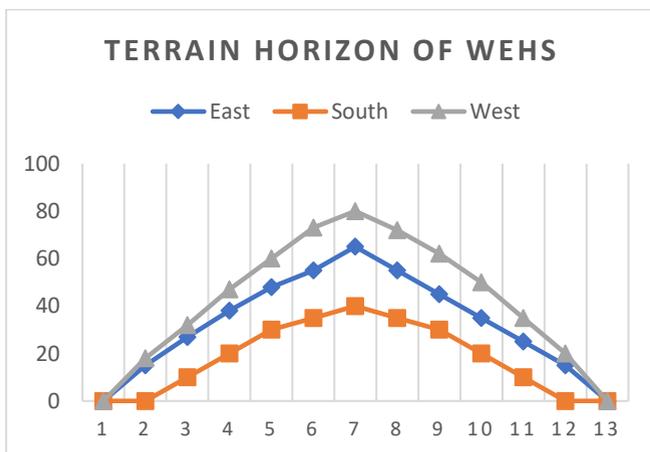
**Fig. 12** Angle of sun over the Study area

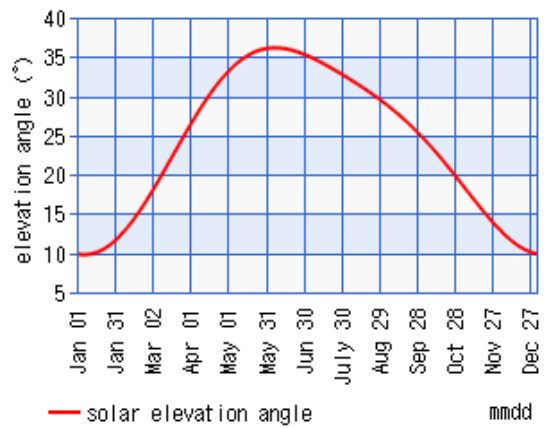
**Fig. 13** Solar elevation angle

Angle of the sun over the study area known as zenith angle is also important in determining the angle of PV panels while installation. It is determined by drawing a vertical perpendicular component between the sun and the study area. Zenith angle of the sun is inversely proportional to the length of the day. Zenith angle of the sun is more in winters when days are short. Very low zenith angle of the sun during summer solstice has been observed in the fig. 14 whereas at vertical equinox it is high.

Global solar irradiations also play important role in changing the electric potential generation capacity of experimental setup. These are a collection of various radiations emitted by the sun and directed at the earth's horizontal surface. These radiations are the total of several types of radiations, including direct, diffuse, and ground reflected radiations. The impact of Global solar irradiations is presented in the Fig. 14 of study area [32]

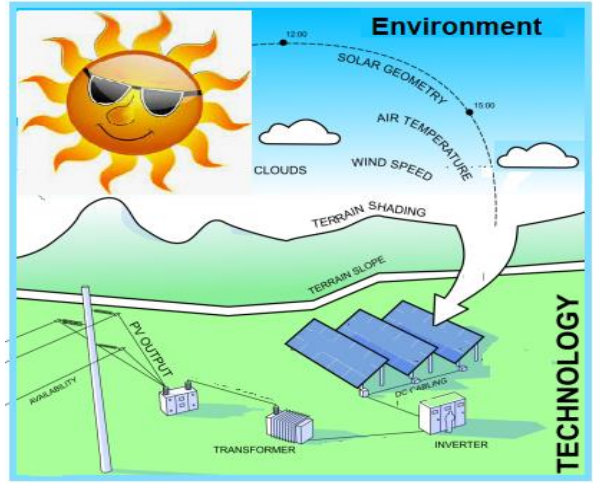

**Fig. 14** Impact of Global solar irradiations

Global solar irradiations are at maximum level in the month of May, and at its minimum level usually in the month of January. An increase in the global irradiations increases the electric power generation. Fig. 6 also presents that more power is generated in May and very less amount of electric power is generated in January.

#### CO2 Emission Reduction
Carbon-di-oxide adds about 80% of all greenhouse gases causing global warming [34]. Carbon emission is increasing day by day causing major changes in the climate but also increasing demand of energy. Use of technologies in producing renewable energy also facilitate to reduce Carbon emission as compared to other sources of energy. These technologies produce less carbon dioxide as compared to other natural sources as presented in Table 7.

**TABLE 7 Renewable energy sources [34]**

| Energy source | Fossil Fuels | Solar | Wind | Hydro | Biogas |
|---|---|---|---|---|---|
| Emission of Carbon (g/kW) | 500% | 95% | 9.2% | 11% | 10% |

Almost $6.9 \times 10^{-4}$ metric tons of carbon di oxide is emitted by generation of 1 kWh of electric energy [34]. Therefor generation of 3,867,502 kWh of solar energy will cause 3,867,502 x 6.9 x $10^{-4} \approx 2,320.5012$ metric tons of $CO_2$ reduction which indicates that PV panel installation of electricity production is only low cost but also environmentally friendly.

## CONCLUSION

The aim of this study is to fulfill the electricity demand of consumer by producing energy locally using PV panels installed in certain area. In this article, FL technique has been applied to determine the total regional need of electricity. Blockchain technology has been used to transmit excessive energy to consumers through electric grid station. Our proposed system contains smart meters that collect data of electricity consumption from every house. Electricity production data is recorded by the digitalizer installed on rooftop. Transmitter has been used to transmit data have been used to transmit recorded parameters to client servers to train local data models. These local models are transferred to the central server for global model training that determines the future need of electricity in the region. After determining energy requirements of the upcoming year, more PV panels might be installed, and more house could be digitalized for producing more electricity to meet demand or trade the excessive energy produced to regional grid station for earning money. We concluded to a point that profit rate is high in the first three months of year as consumption is much less than the production but in summer season, this difference becomes smaller as more electricity is consumed due to rise in regional temperature. This article shows that 9.2 % of total generated energy is used for consumption and the rest of the energy is traded to regional or national grid in adjacent areas.

## Abbreviations

RESs: Renewable Energy Sources; FL: Federated Learning; VCG: Vickrey–Clark–Groves; GRU: Gated Recurrent Unit; ToU: Time of Use; FedGrid: Federated Grid; SC: Smart Contract; WEHS: Wapda Employees Housing Society; PV: Photovoltaic

## Supplementary Information

### Authors' contributions

Muhammad Shoaib Farooq and Rabia Tehseen worked on the conception and design of the paper. Azeen Ahmed Hayat also drafted the paper. Muhammad Shoaib Farooq and Azeen Ahmed Hayat conducted the review and data collection. Rabia Tehseen and Muhammad Shoaib Farooq comprehensively analyzed the full text of the studies. While Azeen Ahmed Hayat validated and verified this research work outcome. Muhammad Shoaib Farooq also refined the concepts and proofread the paper as well. All authors read and approved the final manuscript.

### Authors' information

**Muhammad Shoaib Farooq** is working as Professor at Department of Computer Science at University of Management and Technology, Lahore Pakistan. His research interests include Machine Learning, Distributed Systems, and Data Science.

**Rabia Tehseen** is working as Assistant Professor in Department of Computer Science at University of Central Punjab, Lahore Pakistan. Her research interests include Federated Learning, Machine Learning and Data Science.

**Uzma Omer** is working as Lecturer in Division of Computer Science at University of Education, Lahore Pakistan. Her research interests include Information sciences and Artificial Intelligence.

**Azeen Ahmed Hayat** is currently a Masters Student in University of the Punjab, Lahore. Her research mainly focusses on electric systems.

### Funding

No funding available for this research.

### Data availability statement

The datasets used and/or analyzed during the current study are available from the corresponding author on reasonable request.

## Declarations

### Ethics approval and consent to participate

Not applicable

### Consent for publication

Not applicable

### Competing interests

The authors declare that they have no competing interests.

### Author details

[1]Department of Computer Science, University of Management and Technology, Lahore, Pakistan; [2]Department of Computer Science, University of Central Punjab, Lahore, Pakistan; [3]Department of Computer Science, University of Education, Lahore; [4]Institute of Electrical, Electronics and Computer Engineering, University of the Punjab Lahore, Pakistan